%% file: paper.tex
\documentclass[twoside]{article}
\usepackage{algorithm}
\usepackage{algorithmic}

\usepackage[accepted]{aistats2014}
\input{preamble}

\pgfrealjobname{paper} 

\begin{document}

%

%
\twocolumn[

\aistatstitle{Local Gaussian Regression}



\aistatsauthor{ Franziska Meier$^1$ \And Philipp Hennig$^2$ \And
  Stefan Schaal$^{1,2}$ }
\vspace{0.1cm}
\aistatsaddress{ $^{1}$Computational Learning and Motor Control Lab,
  University of Southern California, Los Angeles  \\
  $^{2}$Max-Planck-Institute for Intelligent Systems, T{\"u}bingen, Germany } ]

\begin{abstract}
  Locally weighted regression was created as a nonparametric learning
  method that is computationally efficient, can learn from very large
  amounts of data and add data incrementally. An interesting feature
  of locally weighted regression is that it can work with spatially
  varying length scales, a beneficial property, for instance, in
  control problems. However, it does not provide a generative model
  for function values and requires training and test data to be
  generated identically, independently. Gaussian (process) regression,
  on the other hand, provides a fully generative model without
  significant formal requirements on the distribution of training
  data, but has much higher computational cost and usually works with
  one global scale per input dimension. Using a localising function
  basis and approximate inference techniques, we take Gaussian
  (process) regression to increasingly localised properties and
  toward the same computational complexity class as locally weighted
  regression.
\end{abstract}

\section{Introduction}
Besides expressivity and sample efficiency, computational cost is a
crucial design criterion for machine learning algorithms in real-time
settings, such as control problems. An example is the problem of
building a model for robot dynamics: The sensors in a robot's limbs
can produce thousands of datapoints per second, quickly amassing a
local coverage of the input domain. In such settings, fast local
learning and generalization can be more important than a globally
optimized model. A learning method should rapidly produce a good model
from the large number $N$ of datapoints, using a comparably small
number $M$ of parameters.

Locally weighted regression (LWR) \cite{schaal1998constructive}
makes use of the popular and well-studied idea of local learning 
\cite{bottou1992local} to address the task of compressing large
amounts of data into a small number of parameters. In the spirit of a Taylor expansion,
the idea of LWR is that simple models with few parameters may locally
be precise, while it may be difficult to find good nonlinear features
to capture the entire function globally -- lots of good local models
may form a good global one.  

The key to LWRs low computational cost (linear, $\mathcal{O}(NM)$) is that each
local model is trained independently. The resulting speed has made LWR
popular in robot learning. The downside is that LWR requires several
tuning parameters, whose optimal values can be highly data
dependent. This is at least partly a result of the strongly localized
training strategy, which does not allow models to `coordinate', or
to benefit from other local models in their vicinity.

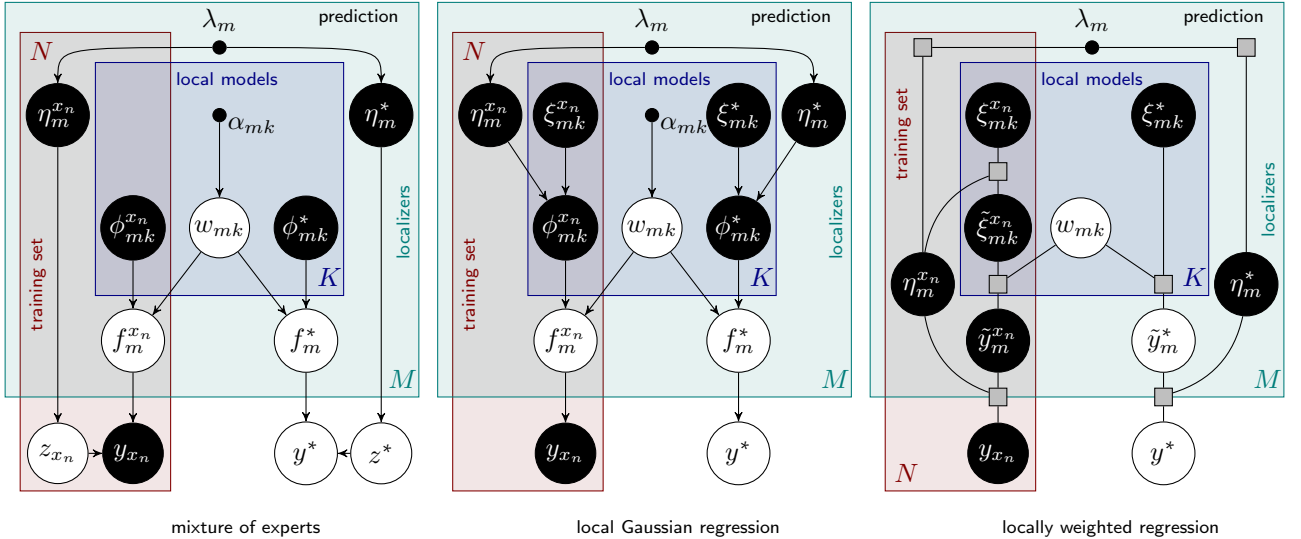
\begin{figure*}[ht]
\centering
   \beginpgfgraphicnamed{GraphicalModels-external}%
   \input{GraphicalModels.tikz}%
   \endpgfgraphicnamed%
 
\caption{\label{fig:graphs}\small{ Generative model for mixture of
    experts (ME) (left), local Gaussian regression (LGR) (middle), and
    factor graph for locally weighted regression (LWR) (right).  LWR
    assumes a fixed contribution of each local model to each
    observation. So a V-structure only exists \emph{within} each
    $m$-plate, making inference cheaper. In ME, contributions of
    experts to each observations are normalized, coupling all the
    experts and making it necessary to update responsibilities in the
    {\sc e}-step in a global manner.  In local Gaussian regression, the
    contribution of each model to observations is uncertain (similar
    to ME), so the model is more densely connected than LWR. However,
    LWR is not a generative model for the data: It treats training and
    test data in different ways.}}
\end{figure*}

Here, we explore a probabilistic alternative to LWR that alleviates
the need for parameter tuning, but retains potential for fast
training. An initial candidate could be the mixture of experts model
(ME) \cite{jacobs1991adaptive}. Indeed, it has been argued
\cite{schaal1998constructive}, that LWR can be thought of as a mixture
of experts model in which experts are trained independently of each
other. The advantage of ME is that it comes with a full generative
model \cite{waterhouse1996bayesian, hannah2011dirichlet} allowing for
principled learning of parameters and expert allocations, while LWR
does not (see Figure~\ref{fig:graphs}). However, in this work we
emulate LWRs assumption that the data is a continous function as
opposed to a mixture of linear models, for instance. When the
underlying data is not assumed to be generated by a mixture model,
utilizing ME to fit the data makes inference (unneccesarily)
complicated. Hence, we are interested in a probabilistic model that
captures the best of both worlds without making the mixture
assumption: A generative model that has the ability to localize
computations to speed up learning.
%
%
%

%
Our proposed solution is \emph{local Gaussian regression} (LGR), a
linear regression model that explicitly encodes localisation.
Starting from the well-known probabilistic, Gaussian formulation of
least-squares generalized linear regression, we first re-interpret the
well known radial basis feature functions as localisers of constant
feature functions. This allows us to introduce more expressive features,
capturing the idea of local models within the Gaussian regression
framework.  In its exact form, this model has the cubic cost in $M$
typical of Gaussian models, arising because observations induce
correlation between all local models in the posterior. To decouple the
local models, we propose a variational approximation that gives
essentially linear cost in the number of models $M$. The core of this
work revolves around fitting the model parameters using maximum
likelihood (Section \ref{sec:learn-local-bayes}).  As a final note, we
show how the novel feature function representation allows us to readily
extend our probabilistic model to a local nonparametric formulation
(Section \ref{sec:nonp-extens-i}).

Previous work on probabilistic formulations of local
regression \cite{Ting:2008uc,Edakunni:2007uf} has been focussed on
bottom-up constructions, trying to find generative models for one
local model at a time. 
To our knowledge, this is the first top-down approach,
starting from a globally optimal training procedure, to find
approximations giving a localized regression algorithm similar in
spirit to LWR.
\section{Background}
\label{sec:background}
Both LWR and Gaussian regression have been studied extensively before,
so we only give brief introductions here. Generalized linear
regression maps weights $\vec{w} \in \mathbb{R}^{F}$ to the nonlinear
function $f:\Re^D\to\Re$ via $F$ feature functions
$\phi_i(x):\Re^D\to\Re$:
\begin{equation}
  \label{eq:3}
  f(x) = \sum_{i=1}^F \phi_i(x) w_{i} = \vec{\phi}\Trans\vec{w}.
\end{equation}
Using the feature matrix $\Phi \in \Re^{N \times F}$ whose
elements are $\Phi_{nf} = \phi_f(x_n)$, the function values at
$N$ locations $\vec{x}_n\in\Re^D$, subsumed in the matrix $X\in
\Re^{N\times D}$, are $\vec{f}(X)=\Phi \vec{w}$.

\paragraph{Locally weighted regression (LWR)}
\label{sec:locally-weight-regr}

trains $M$ \emph{local} regression models. We will assume each local
model has $K$ local feature functions $\xi_{mk}(x)$, so that the
$m$-th model's prediction at $x$ is
\begin{equation}
  \label{eq:7}
  f_{m}(x) = \sum_{k=1} ^K \xi_{mk}(x) w_{mk} = \vec{\xi}_m(x)\vec{w}_m
\end{equation}
$K=2$ and $\xi_{m1}(x)=1, \xi_{m2}(x) = (x-c_m)$ gives a linear model
around $c_m$. The models are localized by a non-negative, symmetric
and integrable weighting $\eta_m(x)$, typically the radial basis
function
\begin{align}
  \label{eq:8}
  \eta_m(x) &= \exp\left[-\frac{(x - c_m)^2}{2\lambda_m ^2} \right],
  \quad \text{ or, for $\x \in\Re^D$,} \\
  \eta_m(\x) &= \exp\left[-\frac{1}{2}(\x -
    \vec{c}_m)\Lambda^{-1}(\x-\vec{c}_m)\Trans \right]. 
\end{align}
with center $c_m$ and length scale $\lambda$ or positive definite
metric $\Lambda$. In LWR, each local model is trained independently of
the others, by minimizing a quadratic loss
\begin{align}
  \label{eq:9}
  \L(\w) &= \sum_{n=1} ^N \sum_{m=1}^M \eta_m(x_n)(y_n - \vec{\xi}_m(x_n)
  \vec{w}_m)^2 \\
  & = \sum_{n=1} ^N \sum_{m=1}^M \left(\sqrt{\eta_m(x_n)}y_n - 
    \sqrt{\eta_m(x_n)} \vec{\xi}_m(x_n)\vec{w}_m\right)^2 \notag
\end{align}
over the $N$ observations $(y_n,x_n)$. At test time, local predictions
(\ref{eq:7}) are combined into a joint prediction at input $x$ as a
normalised weighted sum
\begin{align}
  \label{eq:6}
  f(x) &= \frac{\sum_m \eta_m(x) f_m(x)}{\sum_{m'} \eta_{m'}(x)} = \sum_{m,k}
  \phi_{mk}(x) w_{mk} \\ 
  \intertext{with the features} 
  \phi_{mk}(x) &= \frac{\eta_m(x) \xi_{mk}(x)}{\sum_{m'}\eta_{m'}(x)}.
\end{align}
An important observation is that the objective (\ref{eq:9}) cannot be
interpreted as least-squares estimation of the linear model $\Phi$
from Eq.~(\ref{eq:6}). Eq.~(\ref{eq:9}) is effectively training $M$
linear models with $\phi_{mk}(x_n)=\sqrt{\eta_m(x_n)}\xi_{mk}(x_n)$ on
$M$ \emph{separate} datasets $y_m(x_n) = \sqrt{\eta_m(x_n)}y_n$, but
that model differs from the one in Equation (\ref{eq:6}). Thus, LWR
can not be cast probabilistically as one generative model for training
and test data simultaneously. The factor graph in Figure
\ref{fig:graphs}, right, shows this broken symmetry.

Training LWR is linear in $N$ and $M$, and cubic in $K$, since it
involves a least-squares problem in the $K$ weights $\vec{w}_m$. But
this low cost profile also means absence of mixing terms between the
$M$ models, thus no `coordination' between the local models. One
problem this causes is that model-specific hyperparameters are not
sufficiently identified and tend to over-fit. To pick one example
among the hyperparameters: when learning the length scale parameters
$\lambda_m$, there is no sense of global improvement. Each local model
focusses only on fitting data points within a $\lambda_m$-neighborhood
around $c_m$. So the optimal choice is actually $\lambda\to 0$, which
gives a useless regression model. To counteract this behaviour,
implementations of LWR use regularisation of $\lambda$, but of
course this introduces free parameters, whose best values are not
obvious. Most works in LWR address this problem via leave-one-out
cross-validation, which works well if the data is sparse. But when the
data is dense, like in robotics, the kernel tends to shrink to only
`activate' points close by the left-out data point.
\paragraph{Gaussian regression}
\label{sec:gaussian-regression}

\cite[\textsection 2]{RasmussenWilliams} is a probabilistic framework
for inference on the weights $\vec{w}$ given the features
$\bphi$. Using a Gaussian prior $p(\w)=\N(\w;\bmu_0,\Sigma_0)$ and a
Gaussian likelihood $p(\y\g \bphi,\w) = \N(\vec{y};\vec{\phi}\Trans
\vec{w},\beta^{-1}\Id)$, the posterior distribution  is itself
Gaussian:
\begin{align}
  \label{eq:4}
  p(\w\g \y,\bphi)  &= \N\left(\w; \bmu_N, \Sigma_N\right) \quad \text{with}\\
  \bmu_N &=  (\Sigma^{-1}_0 +\beta\Phi\Trans\Phi)^{-1} (\beta\Phi\Trans \y -
  \Sigma_0 ^{-1} \bmu_0)\\
  \Sigma_N &= (\Sigma_0^{-1} + \beta\Phi\Trans\Phi)^{-1} 
\end{align}
The mean of this Gaussian posterior is identical with the
$\Sigma$-regularised least-squares estimator for $\w$, so it could
alternatively be derived as the minimiser of the loss function
\begin{equation}
  \label{eq:10}
  \L(\w) = \w\Trans\Sigma^{-1}\w + \sum_{n=1} ^N (y_n - \bphi(x_n) \w)^2.
\end{equation}
But the probabilistic interpretation of Equation (\ref{eq:4}) has
additional value over (\ref{eq:10}) because it is a generative model
for all (training and test) data points $y$, which can be used to
learn hyperparameters of the feature functions. The prediction for
$f(x_*)$ with features $\bphi(x_*)=:\bphi_*$ is also Gaussian, with
\begin{equation}
  \label{eq:5}
  p(f(x_*)\g \y,\bphi) = \N(f(x_*); \bphi_*\bmu_N, \bphi_* \Sigma_N \bphi_* \Trans)
\end{equation}
As is widely known, this framework can be extended to the
nonparametric case by a limit which replaces all inner products
$\bphi(x_1)\Sigma_0\bphi(x_2) \Trans$ with a Mercer (positive
semi-definite) kernel function $k(x_1,x_2)$. The corresponding priors
are Gaussian processes. This generalisation will only play a role in
Section \ref{sec:nonp-extens-i} of this paper, but the direct
connection between Gaussian regression and the elegant theory of
Gaussian processes is often cited in favour of this framework. Its
main downside, relative to LWR, is computational cost: Calculating the
posterior (\ref{eq:5}) requires solving the least-squares problem for
all $F$ parameters $\w$ jointly, by inverting the Gram matrix
$(\Sigma_0^{-1} + \beta \Phi\Trans\Phi)$. In the general case, this
inversion requires $\O(F^3)$ operations. The point of the construction
in the following sections is to use approximations to lower the
computation cost of this operation such that the resulting algorithm
is comparable in cost to LWR, while retaining the probabilistic
interpretation, and the modelling robustness of the full Gaussian
model.

The Gaussian regression framework puts virtually no limit on the
choice of basis functions $\phi$, even allowing discontinuous and
unbounded functions, but the radial basis function (RBF,
aka.~Gaussian, square-exponential) features from Equation
(\ref{eq:8}), $\bphi_i(x)=\vec{\eta}_i(x)$, (for $F=M$) enjoy
particular popularity for various reasons, including algebraic
conveniences and the fact that their associated reproducing kernel
Hilbert space lies dense in the space of continuous functions
\cite{micchelli2006universal}. A downside of this covariance function
is that it uses one global length scale for the entire input
domain. There are some special kernels of locally varying regularity
\cite{gibbs1997bayesian}, and mixture descriptions offer a more
discretely varying model class \cite{rasmussen2002infinite}. Both,
however, require computationally demanding training.
\begin{figure*}[htb]
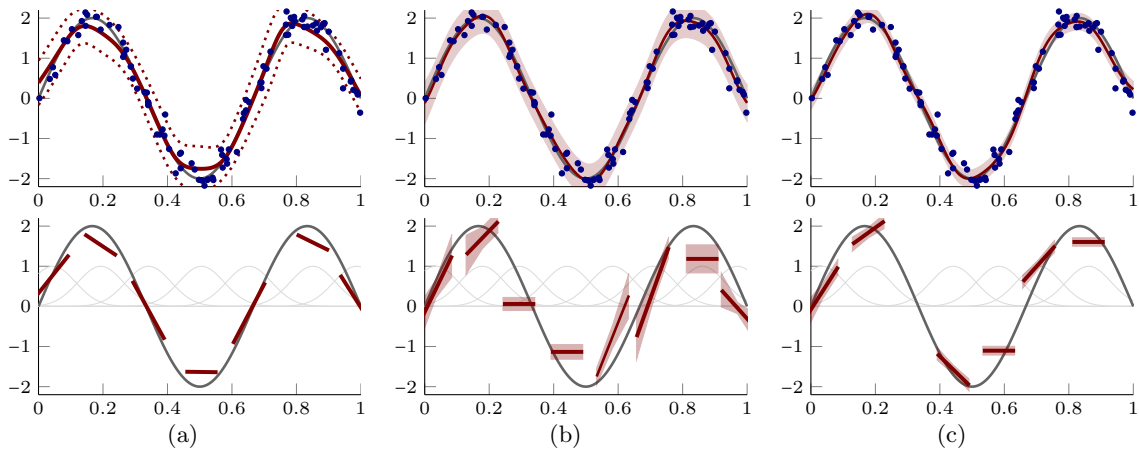

  \centering
  \scriptsize
  \subfigure[]{%
   \beginpgfgraphicnamed{lwr_sine-external}%
   \input{lwr_sine.tikz}%
   \endpgfgraphicnamed%
 }
  \subfigure[]{%
   \beginpgfgraphicnamed{plr_sine-external}%
   \input{plr_sine.tikz}%
   \endpgfgraphicnamed%
 }
  \subfigure[]{%
   \beginpgfgraphicnamed{approx_plr_sine-external}%
   \input{approx_plr_sine.tikz}%
   \endpgfgraphicnamed%
 }
  \caption{\small{Noisy data drawn from sine function learned by (a) LWR (b)
    exact and (c) approximate local Gaussian regression.  In LWR,
    local models do not know of each other and thus fit tangential
    lines around the center of each local model. In local Gaussian
    regression local models are correlated.  They work together to fit
    the function. In the approximate version (c) this correlation is
    reduced, but not completely gone.}}
  \label{fig:lr_illu}
\end{figure*}
\section{Local Parametric Gaussian Regression}
\label{sec:local-bayes-regr}

In Gaussian regression with RBF features as described above, without
changing anything, the features $\phi_m(x)=\eta_m(x)$ can be
interpreted as $M$ constant function models $\xi_m(x)=1$, localised by
the RBF function, $\bphi(x) =[ \xi_1(x) \eta_1(x), \dots, \xi_M (x)
\eta_M(x)]$. This representation extends to more elaborate local
models $\bxi(x)$. For example $\bxi(x) = x$ gives a local weighted
regression with linear local models. Extending to $M$ local models
consisting of $K$ parameters each, feature function $\phi^n _{mk}$
combines the $k^{th}$ component of the local model $\xi_{km}(x_n)$,
localised by the $m$-th weighting function $\eta_m(x)$
\begin{equation}
  \label{eq:2}
  \phi^n _{mk} := \phi _{mk}(x_n) =  \eta_m(x_n) \xi_{km}(x_n).
\end{equation}
For these features, treating $mk$ as the index set of a vector with
$MK$ elements, the results from Equations (\ref{eq:4}) and
(\ref{eq:5}) apply, giving a localised linear Gaussian regression
algorithm.

The choice of local parametric model is essentially arbitrary. But
regular features are an obvious choice. A scalar polynomial model of
order $K-1$ arises from $\xi_{km}(x_n) = (x_n-c_m)^{k-1}$. Local
linear regression in a $K$-dimensional input space takes the form
$\xi_{km}(\x_n) = x_{nk}-c_m$. We will adopt this latter model in the
remainder, because it is also the most widely used model in LWR.  An
illustration of LWR regression and the proposed local Gaussian
regression is given in Figure~\ref{fig:lr_illu}. 

Since it will become necessary to prune out unnecessary parts of the
model, we adopt the classic idea of automatic relevance determination
\cite{neal1996bayesian,Tipping:2001wb} using a factorizing prior
\begin{align}
  \label{eq:17}
   p(\vec{w} | \mat{A} ) =& \prod_{m=1}^M \N(\w_m;0, A_m ^{-1}) \quad
   \text{with} \\ A_m &= \diag(\alpha_{m1}, \dots, \alpha_{mK}).
 \end{align}
 So every component $k$ of each local model $m$ has its own precision,
 and can thus be pruned out by setting $\alpha_{mk}\to\infty$. Section
 \ref{sec:learn-local-bayes} assumes a fixed number $M$ of local
 models with fixed centers $c_m$. Thus the parameters are
 $\vec{\theta} = \{\beta, \{\alpha_{mk}\}, \{\lambda_{mk}\}\}$. We
 propose an approximation for estimating these parameters. Section
 \ref{sec:incr-adding-local} then describes placing the local models
 incrementally to adapt $M$ and $c_m$.

\subsection{Learning in Local Bayesian Linear Regression}
\label{sec:learn-local-bayes}
%
Exact inference in Gaussian regression with localised feature
functions comes with the cubic cost of its nonlocal origins. However,
because of the localised feature functions, correlation between far
away local models is approximately $0$, hence inference is approximately
independent between local models. In this section we aim to make use of
this ``almost independence'' to derive a localised approximate inference
scheme of low cost, similar in spirit to LWR. To arrive at this
localised learning algorithm we first introduce a latent variable $f_{nm}$
for each local model $m$ and data point $x_n$, similar to
probabilistic backfitting \cite{dsouza2004}. Intuitively, the
$\vec{f}$ form approximate targets, one for each local model, against
which the local regression parameters fit (see also Figure~\ref{fig:graphs}, middle).

This modified model motivates a factorizing variational bound
constructed in Section \ref{sec:variational-bound}, rendering the
the local models computationally independent, which allows for fast
approximate inference in the local Gaussian model. Hyperparameters
will be learnt by approximate maximum likelihood
(\ref{sec:optim-hyperp-}), i.e.~iterating between constructing a bound
$q(\vec{z}\g\vec\theta)$ on the posterior over variables $\vec{z}$
given current parameter estimates $\vec\theta$ and optimising $q$ with
respect to $\vec\theta$.

\subsubsection{Variational Bound}
\label{sec:variational-bound}

The complete data likelihood of the modified model is
\begin{align}
  p(\vec{y},& \vec{f}, \vec{w} \g \Phi, \theta) = \prod_{n=1}^N
    p(y_n \g \vec{f}_n, \beta_y) \\ 
   & \prod_{n=1}^N \prod_{m=1}^M p(f_{nm} \g \bphi_m^n \vec{w}_m,
   \beta_{fm}) \prod_{m=1}^M p(\vec{w}_m\g A_m) \notag
\end{align}
(c.f.~Figure~\ref{fig:graphs}, left). Our Gaussian model involves the
latent variables $\w$ and $\vec{f}$, the precision $\beta$ and the model
parameters $\lambda_m,c_m$. We treat $\w$ and $\vec{f}$ as probabilistic
variables and estimate $\theta=\{\beta,\vec{\lambda},\vec{c}\}$. On
$\w,\vec{f}$, we construct a variational bound $q(\w,\vec{f})$ imposing
factorisation $q(\vec{w},\vec{f}) = q(\vec{w}) q(\vec{f})$. The
variational free energy is a lower bound on the log evidence for the
observations $y$:
\begin{align}
  \label{eq:13}
  \log p(\y\g \theta) \geq \int q(\w, \vec{f}) \log\frac{p(\y,\w, \vec{f}
    \g\theta)}{q(\w, \vec{f})}.
\end{align}
This bound is maximized by the $q(\w, \vec{f})$ minimizing the
relative entropy $D_{\KL}[q(\w, \vec{f})\|p(\w, \vec{f} \g\y,\theta)]$,
the distribution for which $ \log{ q(\vec{w})} = \Exp_{\vec{f}}{[\log
  p( \vec{y} \g \vec{f}, \vec{w})p(\vec{w}, \vec{f})]}$ and
$\log{q(\vec{f})} = \Exp_{\vec{w}}{[\log p( \vec{y} \g \vec{f},
  \vec{w})p(\vec{w}, \vec{f})]}$. It is relatively straightforward to
show
(e.g.~\cite{titsias09:_variat_learn_induc_variab_spars_gauss_proces})
that these distributions are Gaussian in both $\w$ and $\vec{f}$.The
approximation on $\w$ is
\begin{align}
  \log q(\vec{w}) &= \myExp{\vec{f}}{ \log{p(\vec{f}_n \g \bphi(x_n),
      \vec{w})} + \log{p(\vec{w} \g A)} } \notag \\
   &= \log\prod_{m=1}^M
     \N(\w_m;\vec{\mu}_{w_m},\Sigma_{w_m}) 
\end{align}
where
\begin{align}
\label{eq:27}
  \Sigma_{w_m} &= \left(\beta_{fm} \sum_{n=1}^N \bphi_m^n
  {\bphi_m^n}^T + A_m\right)^{-1} \quad
  \in \mathbb{R}^{K \times K} \\
  \label{eq:28}
  \vec{\mu}_{w_m} &=  \beta_{fm} \Sigma_{w_m} \left(\sum_{n=1}^N
    \bphi_m^n  \myExp{}{f_{nm}}\right) \quad\, \in
    \mathbb{R}^{K \times 1}
\end{align}
The posterior update equations for the weights are local: each of the
local models updates its parameters independently. This comes at the
cost of having to update the belief over the variables $f_{nm}$, which
achieves a coupling between the local models. The Gaussian variational
bound on $\vec{f}$ is
\begin{align}
  \label{eq:14}
  \log{q(\vec{f}_n} ) &= \myExp{\vec{w}}{\log{p(y_n \g
      \vec{f}_n,\beta_y)} + \log{p(\vec{f}_n \g \bphi_m^n,
      \vec{w})} } \notag\\
  & = \N(\vec{f}_n;\vec{\mu}_{fn}, \Sigma_{f})
\end{align}
where
\begin{align}
  \label{eq:16}
  \Sigma_f &= \mat{B}^{-1} - \mat{B}^{-1} \vec{1} (
  \beta_y^{-1} + \vec{1}^T \mat{B}^{-1} \vec{1})^{-1} \vec{1}^T
  \mat{B}^{-1} \notag \\
  &= \mat{B}^{-1} - \frac{\mat{B}^{-1} \vec{1} \vec{1}^T
    \mat{B}^{-1} }{ \beta_y^{-1} +
    \vec{1}^T \mat{B}^{-1} \vec{1}}\\
  \label{eq:15}
  \vec{\mu}_{fn} &= \sum_{m=1}^M \myExp{\vec{w}}{\w_m^T} \bphi(x_n) +\\
  & \frac{1}{\beta_y ^{-1} + \vec{1}\Trans \mat{B}^{-1}\vec{1}}
  \mat{B}^{-1} \vec{1} \left(y_n - \sum_{m=1}^M \myExp{}{\vec{w}_m}^T
    \bphi_m^n \right) \notag
\end{align}
where $\mu_{fn} \in \mathbb{R}^{M}$ and using $\mat{B} =
\diag{(\beta_{f1}, \dots, \beta_{fM})}$. These updates can be
performed in $\mathcal{O} (MK)$. Inspecting Equations (\ref{eq:15})
and (\ref{eq:28}), one can see that the optimal assignment of all
$\bmu_w$ actually amounts to a joint linear problem of size $MK\times
MK$, which could also be solved as a linear program. However, this
would come at additional computational cost.

\subsubsection{Optimizing Hyperparameters}
\label{sec:optim-hyperp-}

We set the model parameters $\theta = \{\beta_y, \{ \beta_{fm},
\lambda_m\}_{m=1}^M , \{ \alpha_{mk} \} \}$ to maximize the expected
complete log likelihood under the variational bound,
\begin{align}
\Exp_{\vec{f},\w} [\log p&(\vec{y}, \vec{f}, \vec{w} \g \Phi,
    \theta)]  = \\ \notag
  &\quad\Exp_{\vec{f},\vec{w}}\bigg\{ \sum_{n=1}^N \bigg[
  \log \N\left(y_n;\sum_{m=1}^M f_{nm}, \beta^{-1} _y\right)\\ \notag
       &\qquad+ \sum_{m=1}^M 
     \log \N(f_{nm};\vec{w}_m^T \bphi_m^n,\beta^{-1} _{fm}) \bigg]\\
     &\qquad + \sum_{m=1}^M \log \N(\vec{w}_m;0, A^{-1}_m) \bigg\} \notag
\end{align}
Setting the gradient of this expression to zero leads to the following
update equations for the variances
\begin{align}
\beta_y ^{-1} &= \frac{1}{N} \sum_{n=1}^N (y_n - \vec{1} \vec{\mu}_{fn})^2
+ \vec{1}^T \Sigma_{f} \vec{1} \label{eq:18} \\
\beta^{-1} _{fm} &= \frac{1}{N} \sum_{n=1}^N\left[ (\mu_{fnm} -
  \vec{\mu}_{w_m} \bphi_m^n)^2 + {\bphi_m ^n}^T \Sigma_{w_m}
  \bphi_m^n \right] + \sigma_{fm}^2 \label{eq:betafm}\\
\alpha_{mk}^{-1} & = \mu_{w_{mk}}^2 + \Sigma_{w,kk} \label{eq:alphamk}
\end{align}

The gradient with respect to the scales of each local model is
completely localized
\begin{align}
  &\frac{ \de \myExp{ \vec{f},\vec{w} }{ \log{ p(\vec{y} , \vec{f},
        \vec{w} \g \Phi, \theta) } } } {\de \log{\lambda_{mk}}} \notag \\
  &\quad =\frac{\de \myExp{\vec{f},\w} { \sum_{n=1}^N \N(f_{nm}; \w_m^T
      \bphi_m(x_n),\beta_{fm}^{-1} ) } }{\de \lambda_{mk}} \label{eq:lambdamk}
\end{align}
We use gradient ascent to optimize the length scales $\lambda_{mk}$. All
necessary equations are of low cost and, with the exception of the
variance $\frac{1}{\beta_y}$, all hyper-parameter updates are solved
independently for each local model, similar to LWR. In contrast to
LWR, however, these local updates do not cause a shrinking in the
length scales: In LWR, both inputs and outputs are weighted by the
localizing function, thus reducing the length scale improves the
fit. The localization of Equation~(\ref{eq:2}) only affects the
influence of regression model $m$, but the targets still need to be
fit accordingly. Shrinking of local models only happens if it actually
improves the fit against the unweighted targets $f_{nm}$.

\subsubsection{Prediction}
Prediction at a test point $x_{*}$ arise from marginalizing over both
$\w$ and $\vec{f}$, using
\begin{align}
  \label{eq:20}
  &\int \N(y_{*};\vec{1}^T \vec{f}_{*},\beta_y^{-1})
  \N(\vec{f}_{*};\mat{W}^T \bphi(x_{*}),\mat{B}^{-1}) d\vec{f_*}  \notag \\
  & \qq = \N(y_{*};\sum_m \vec{w}^T_m \bphi_m^{*},\beta_y^{-1} + \vec{1}^T
    \mat{B}^{-1} \vec{1})
\end{align}
and
\begin{align}
  \label{eq:21}
  \int \N(y_{*};& \vec{w}^T \bphi^{*} ,\beta_y^{-1} + \vec{1}^T
  \mat{B}^{-1} \vec{1})
  \N(\vec{w};\vec{\mu}_w,\mat{\Sigma_{w}}) d\vec{w} \notag \\
  &= \N\left(y_{*};\sum_{m}^M \vec{\mu}_{w_m}^T
    \bphi_m^*, \sigma^2(x^*) \right)
\intertext{where}
\sigma^2(x^*) &=  \beta_y^{-1} + \sum_{m=1}^M \beta_{fm}^{-1} +
    \sum_{m=1}^M {\bphi_m^{*}}^T \Sigma_{w_m} \bphi_m^{*}
\end{align}
which is linear in $M$ and $K$.
\subsection{Incremental Adding of Local Models}
\label{sec:incr-adding-local}
Up to here, the number $M$ and locations $\vec{c}$ of local models
were fixed. An extension analogous to the incremental learning of the
relevance vector machine \cite{quinonero2002incremental} can be used
to iteratively add local models at new, greedily selected locations
$c_{M+1}$. The resulting algorithm starts with one local model, and
per iteration adds one local model in the variational
step. Conversely, existing local models for which all components
$\alpha_{mk} \to \infty$ are pruned out. This works well in practice,
with the caveat that the number local models $M$ can grow fast
initially before the pruning becomes effective. Thus we check for each
selected location $c_{M+1}$ whether any of the existing local models
$c_{1:M}$ produces a localizing weight $\eta_m(c_{M+1}) \ge
w_\text{gen}$, where $w_\text{gen}$ is a parameter between $0$ and $1$
and regulates how many parameters are added. An overview of the final
algorithm is given in \ref{alg:incLGR}.

\begin{algorithm}[t]
\caption{Incremental LGR}\label{alg:incLGR}
\begin{algorithmic}[1]
\REQUIRE $\{ x_n, y_n\}_{n=1}^N$
\COMMENT{initialize first local model}
\STATE $M = 1; c_1 = x_1; C = \{ c_1 \}$
\COMMENT{iterate over all data points}
\FOR{$ n = 2 \dots N$}
\IF{$\eta_m(x_n) < w_\text{gen} ,\forall m = 1, \dots, M$}
\STATE $c_m \gets x_n$
\STATE $C \gets C \cup \{c_m\}, \quad M = M+1$
\ENDIF  
\COMMENT{{\sc E}-Step: equations
  \eqref{eq:27},\eqref{eq:28},\eqref{eq:16},\eqref{eq:15}} 
\STATE $\{\mu_{w_m}, \Sigma_{w_m}, \mu_{fm}, \Sigma_{fm}\}_{m} \gets$
\STATE $\qqq \text{\textsc{e-step}}(\{\beta_y,\beta_{fm}, \{\alpha_{mk},
\lambda_{mk}\}_k \}_m$) 
\COMMENT{{\sc M}-Step: equations
  \eqref{eq:18},\eqref{eq:betafm},\eqref{eq:alphamk},\eqref{eq:lambdamk}} 
\STATE $\{\beta_y,\beta_{fm}, \{\alpha_{mk}, \lambda_{mk}\}_k \}_m \gets $
\STATE $ \qqq \text{\textsc{m-step}} \left(\{\mu_{w_m}, \Sigma_{w_m}, \mu_{fm},
\Sigma_{fm}\}_{m} \right)$ 
\COMMENT{pruning}
\FORALL{$m$}
\IF{$\alpha_{mk} > 1e3 \quad \forall k = 1, \dots, K$ }
\STATE $M \gets M-1; C \gets C \setminus c_m$
\ENDIF
\ENDFOR
\ENDFOR
\end{algorithmic}
\end{algorithm}
\begin{figure*}[htb]
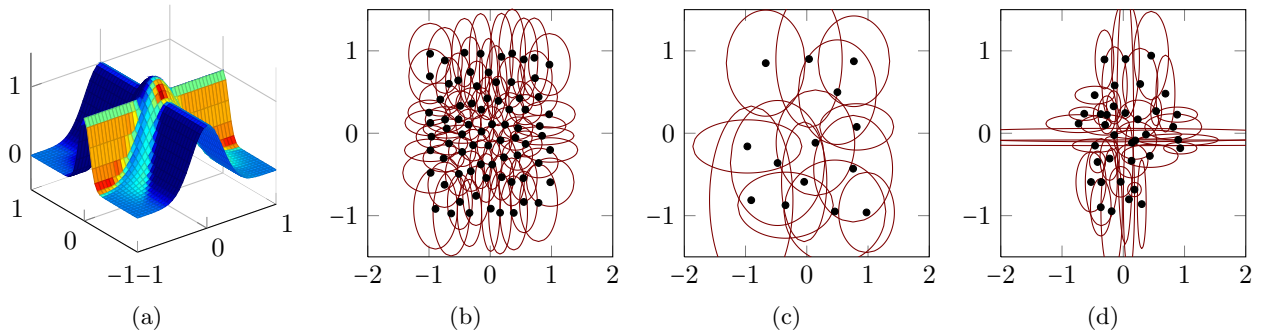

  \centering
  \subfigure[]{%
   \beginpgfgraphicnamed{cross2D_gt_bwa-external}%
   \input{cross2D_gt_bwa.tikz}%
   \endpgfgraphicnamed%
 }
  \subfigure[]{%
   \beginpgfgraphicnamed{cross2D_rfs_bwa-external}%
   \input{cross2D_rfs_bwa.tikz}%
   \endpgfgraphicnamed%
 }
  \subfigure[]{%
   \beginpgfgraphicnamed{cross2D_rfs_lgr-external}%
   \input{cross2D_rfs_lgr.tikz}%
   \endpgfgraphicnamed%
 }
  \subfigure[]{%
   \beginpgfgraphicnamed{cross2D_rfs_me-external}%
   \input{cross2D_rfs_me.tikz}%
   \endpgfgraphicnamed%
 }
 \caption{\small{(a) 2D cross function, local models learnt by  (b) LWPR,
    (c)  LGR and (d) ME }.}
  \label{fig:cross_fun}
\end{figure*}
\section{Extension to Finitely Many Local Nonparametric Models}
\label{sec:nonp-extens-i}
An interesting component of local Gaussian regression is that it
easily extends to a model with finitely many local nonparametric,
Gaussian process models.  Marginalising out the weights $\vec{w}_m$ in
Eq.~(\ref{eq:13}) gives the marginal \vspace{-0.3cm}
\begin{equation}
 p(\vec{y},\vec{f} \g \mat{X}, \theta) = \N(\vec{y}; \vec{1}\Trans\vec{f},
 \beta_y^{-1} \mat{I}_N) \prod_{m=1}^M \N( \vec{f}_m; 0 , C_m)
\end{equation}
where $p(f_m) = \N(\vec{f}_m; 0, C_m)$ is a Gaussian process prior
over function $f_m$ with a (degenerate) covariance function
$C_m=\beta_{fm}^{-1} \mat{I}_N + \bphi_m^T \mat{A}_m^{-1}
\bphi_m$. Replacing the finitely many local features $\xi_m$ with
infinitely many features results in the local nonparametric model with
covariance function $\kappa_m(x,x') = \beta_{fm}^{-1}\delta_{ij} +
\eta_m(x) \hat{\kappa}_m(x,x') \eta_m(x')$. 
\label{sec:experiments}
\begin{table*}[htb]
\caption{Predictive performance on Cross Function}
\label{tab:cross_fun}
\begin{center}
\begin{tabular}{c |c c c | c c c }
  \toprule
  & nMSE w/o LSL & opt $w_\text{gen}$& $\#$ of LMs & nMSE with LSL & opt $w_\text{gen}$ & $\#$ of LMs \\ 
  \midrule
  LWPR   &   0.234 & 0.2 &  13 & 0.0365 & 0.1 & 45.5\\ 
  GLR    &   $\mathbf{0.069}$  & 1.0 & 23.8  &  $\mathbf{0.0137}$ & 0.9 & 23.3 \\ 
  ME    &    0.169  & 0.5 & 21.2  & 0.0313 & 0.4 & 62.2 \\
  \bottomrule
\end{tabular}
\end{center}
\end{table*}
The exact posterior over $\vec{f}_x:=\vec{f}(x)$ is
\begin{equation}
p(\vec{f}_x\g \y) = \N(\vec{f}_x ;\bmu_x ,\Sigma _{xx}) 
\end{equation}
with
\begin{align}
\mat{\Sigma}^{mr} _{xx'} &= \operatorname{cov}(f^m (x),f^r (x'))\\
&=\delta_{mr} k^m _{xx'} - k_{xX} ^m (\sum_s ^M k^s _{XX} + \beta_y ^{-1}
\Id)^{-1}k_{Xx'} ^r
\\
\vec{\mu}^m _x &= k^m _{xX} (\sum_s ^M k^s _{XX} + \beta_y ^{-1} \Id)^{-1} \vec{y}
\end{align}
So computing the posterior covariance $\mat{\Sigma}^{mr} _{xx'}$
requires one inversion of an $N\times N$ matrix. It remains to be seen
to what extend variational bounds on the parametric forms
presented in the previous sections can be transported to a \emph{local
  nonparametric} Gaussian regression framework.
%
%
%
\vspace{-0.2cm}
\input{Experiments}
%
\section{Conclusion}
\label{sec:conclusion}
We have taken a top-down approach to developing a probabilistic
localised regression algorithm: We start with the generative model of
Gaussian generalized linear regression, which amounts to a fully
connected graph and thus has cubic inference cost. To break down the
computational cost of inference, we first introduce the idea of
localised feature functions as local models, which can be extended to
local nonparametric models. In as second step, we
argue that because of the localisation these local models are
approximately independent . We exploit that fact through a variational
approximation that reduced computational complexity to local
computations. Empirical evaluation suggests that LGR
successfully addresses the problem of fitting hyperparameters inherent
in locally weighted regression. A final step left for future work is
to re-formulate our algorithm into an incremental version that can
deal with a continuous stream of incoming data.

\small{\bibliographystyle{unsrt}
\bibliography{regression,bibfile}
}

\end{document}

%% file: preamble.tex
\usepackage{textcase,microtype, marvosym} 

\usepackage{tikz,pgfplots}
\pgfplotsset{compat=newest}
\pgfplotsset{plot coordinates/math parser=false}

\usepackage{amsmath,amssymb,graphicx,MnSymbol}
\usepackage{todonotes}
\usepackage{subfigure}

\definecolor{lred}{RGB}{200,0,0} 
\definecolor{dred}{RGB}{130,0,0} 

\definecolor{dblu}{RGB}{0,0,130} 
\definecolor{dgre}{RGB}{0,130,0} 

\definecolor{dgra}{RGB}{50,50,50}
\definecolor{mgra}{RGB}{100,100,100}
\definecolor{lgra}{RGB}{220,220,220}
\definecolor{MPG}{RGB}{000,125,122}
\definecolor{ora}{RGB}{255,153,51}

\usepackage{booktabs,colonequals}

\usepackage{comment}

\newcommand{\g}{\,|\,} 
\newcommand{\de}{\partial}
 
\newcommand{\Exp}{\mathbb{E}}

\newcommand{\KL}{\text{KL}} 
\renewcommand{\Re}{\mathbb{R}}

\newcommand{\N}{\mathcal{N}}

\renewcommand{\L}{\mathcal{L}}
\newcommand{\Trans}{^{\intercal}}

\newcommand{\diag}{\operatorname{diag}}

\newcommand{\myExp}[2]{\mathbb{E}_{#1}\left[ #2 \right] }

\newcommand{\qq}{\qquad}
\newcommand{\qqq}{\quad\qquad}

\renewcommand{\vec}{\boldsymbol} 
\newcommand{\mat}{}

\renewcommand{\O}{\mathcal{O}} 

\newcommand{\Id}{\vec{I}}

\newcommand{\bmu}{\boldsymbol{\mu}}
\newcommand{\bxi}{\boldsymbol{\xi}}

\newcommand{\bphi}{\boldsymbol{\phi}}

\newcommand{\w}{\vec{w}}
\newcommand{\y}{\vec{y}}
\newcommand{\x}{\vec{x}}

\usetikzlibrary{arrows,shapes,plotmarks}

\tikzset{>=stealth'} 
\tikzstyle{graphnode} = 
   [circle,draw=black,minimum size=22pt,text centered,text
     width=22pt,inner sep=0pt] 
\tikzstyle{var}   =[graphnode,fill=white]
\tikzstyle{obs}   =[graphnode,fill=black,text=white]
\tikzstyle{fac}   =[rectangle,draw=black,fill=black!25,minimum size=5pt]
\tikzstyle{facprior} =[rectangle,draw=black,fill=black,text=white,minimum size=5pt]
\tikzstyle{edge}  =[draw=white,double=black,thick,-]
\tikzstyle{prior} =[rectangle, draw=black, fill=black, minimum size=
5pt, inner sep=0pt]
\tikzstyle{dirprior} = [circle, draw=black, fill=black, minimum
size=5pt, inner sep=0pt]

\DeclareSymbolFont{stmry}{U}{stmry}{m}{n}
\DeclareMathSymbol\leftarrowtriangle\mathrel{stmry}{"5E}
\DeclareMathSymbol\rightarrowtriangle\mathrel{stmry}{"5F}
\renewcommand{\gets}{\operatorname*{\leftarrowtriangle}}
\renewcommand{\to}{\operatorname*{\rightarrowtriangle}}

\newif\iffinal 

\iffinal
 
\else
 
\fi

\newcounter{mycomment}

\newcounter{PHcomment}

%% file: GraphicalModels.tikz
 \begin{tikzpicture}
%
%
%
%

  \begin{scope}[node distance=1.5cm] 
    \draw[draw=dred,fill=dred,fill opacity=0.1] (-1.5,-0.5) rectangle (0.5,5.6); 
    \draw[draw=MPG,fill=MPG,fill opacity=0.1] (-1.7,0.75) rectangle (3.8,6.0); 
    \draw[draw=dblu,fill=dblu,fill opacity=0.1] (-0.5,2.1) rectangle (2.8,5.2); 

    \node[dred,anchor=north west] at (-1.5,5.6) {$N$};
    \node[MPG,anchor=south east] at (3.9,0.75) {$M$};
    \node[dblu,anchor=south east] at (2.9,2.1) {$K$};

    \node[dred,rotate=90,anchor=north] at (-1.5,2.25) {\scriptsize \sf
      training set};
    \node[MPG,rotate=90,anchor=south] at (3.8,3.075) {\scriptsize \sf
      localizers};
    \node[dblu,anchor=north] at (1.25,5.2) {\scriptsize \sf local models};
    \node[anchor=center] at (3,5.8) {\scriptsize \sf prediction};

    \node[obs] at (0,0) (y) {$y_{x_n}$};
    \node[var] at (-1,0) (z) {$z_{x_n}$} edge[->] (y);
    \node[var, above of=y] (f) {$f^{x_n} _{m}$} edge[->](y);
    \node[obs, above of=f] (phi) {$\phi^{x_n} _{mk}$} edge [->] (f);
    \node[obs] at (-1,4.5) (eta) {$\eta_{m} ^{x_n}$} edge [->] (z);

    \node[var, right of=phi,xshift=-0.35cm] (w) {$w_{mk}$} edge [->] (f);
    \node[dirprior,above of=w,yshift=-.0cm] (pa) {} edge[->] (w); 
    \node[anchor=west,inner sep=1pt,yshift=-.2cm] at (pa.north east) {$\alpha_{mk}$};

    \node[obs, right of=w,xshift=-0.35cm] (phis) {$\phi^* _{mk}$};
    \node[var, below of=phis] (fs) {$f^* _{m}$} edge[<-] (w) edge [<-] (phis);
    \node[var, below of=fs]   (ys) {$y^*$} edge[<-](fs);
    \node[obs] at (3.3,4.5) (etas) {$\eta_{m} ^*$};
    \node[var] at (3.3,0) (zs) {$z^*$} edge[<-](etas) edge[->] (ys);

    \node[dirprior,above of=pa,yshift=-.6cm] (lp) {};
    \draw[->] (lp) .. controls (-1,5.4) .. (eta);
    \draw[->] (lp) .. controls (3.3,5.4) .. (etas);
    \node[anchor=south] at (lp.north) {$\lambda_m$};


     \node at (1.5,-1) {\sf\scriptsize mixture of experts};
  \end{scope}

  \begin{scope}[node distance=1.5cm, xshift=5.75cm]
    \draw[draw=dred,fill=dred,fill opacity=0.1] (-1.5,-0.5) rectangle (0.5,5.6); 
    \draw[draw=MPG,fill=MPG,fill opacity=0.1] (-1.7,0.75) rectangle (3.8,6.0); 
    \draw[draw=dblu,fill=dblu,fill opacity=0.1] (-0.5,2.1) rectangle (2.8,5.2); 

    \node[dred,anchor=north west] at (-1.5,5.6) {$N$};
    \node[MPG,anchor=south east] at (3.9,0.75) {$M$};
    \node[dblu,anchor=south east] at (2.9,2.1) {$K$};

    \node[dred,rotate=90,anchor=north] at (-1.5,2.25) {\scriptsize \sf
      training set};
    \node[MPG,rotate=90,anchor=south] at (3.8,3.075) {\scriptsize \sf
      localizers};
    \node[dblu,anchor=north] at (1.25,5.2) {\scriptsize \sf local models};
    \node[anchor=center] at (3,5.8) {\scriptsize \sf prediction};

    \node[obs] at (0,0) (y) {$y_{x_n}$};
    \node[var, above of=y] (f) {$f^{x_n} _{m}$} edge[->](y);
    \node[obs, above of=f] (phi) {$\phi^{x_n} _{mk}$} edge [->] (f);
    \node[obs, above of=phi] (xi) {$\xi_{mk} ^{x_n}$} edge [->] (phi);
    \node[obs, left of=xi,xshift=0.5cm] (eta) {$\eta_{m} ^{x_n}$} edge [->] (phi);
    
    \node[var, right of=phi,xshift=-0.35cm] (w) {$w_{mk}$} edge [->] (f);
    \node[dirprior,above of=w] (pa) {} edge[->] (w); 
    \node[anchor=west,inner sep=1pt,yshift=-0.2cm] at (pa.north east) {$\alpha_{mk}$};

    \node[obs, right of=w,xshift=-0.35cm] (phis) {$\phi^* _{mk}$};
    \node[var, below of=phis] (fs) {$f^* _{m}$} edge[<-] (w) edge [<-] (phis);
    \node[var, below of=fs]   (ys) {$y^*$} edge[<-](fs);
    \node[obs, above of=phis] (xis) {$\xi_{mk} ^*$} edge [->] (phis);
    \node[obs, right of=xis,xshift=-.5cm] (etas) {$\eta_{m} ^*$} edge [->] (phis);

    \node[dirprior,above of=pa,yshift=-.6cm] (lp) {};
    \draw[->] (lp) .. controls (-1,5.4) .. (eta);
    \draw[->] (lp) .. controls (3.3,5.4) .. (etas);
    \node[anchor=south] at (lp.north) {$\lambda_m$};


     \node at (1.5,-1) {\sf\scriptsize local Gaussian regression};
  \end{scope}
  
  \begin{scope}[node distance=0.75cm,xshift=11.5cm]

    \draw[draw=dred,fill=dred,fill opacity=0.1] (-1.5,-0.5) rectangle (0.5,5.6); 
    \draw[draw=MPG,fill=MPG,fill opacity=0.1] (-1.7,0.75) rectangle (3.8,6.0); 
    \draw[draw=dblu,fill=dblu,fill opacity=0.1] (-0.5,2.1) rectangle (2.8,5.2); 

    \node[dred,anchor=south west] at (-1.5,-0.5) {$N$};
    \node[MPG,anchor=south east] at (3.9,0.75) {$M$};
    \node[dblu,anchor=south east] at (2.9,2.1) {$K$};

    \node[dred,rotate=90,anchor=north] at (-1.5,4.25) {\scriptsize \sf
      training set};
    \node[MPG,rotate=90,anchor=south] at (3.8,3.375) {\scriptsize \sf
      localizers};
    \node[dblu,anchor=north] at (1.25,5.2) {\scriptsize \sf local models};
    \node[anchor=center] at (3,5.8) {\scriptsize \sf prediction};

    \node[obs] at (0,0) (y) {$y_{x_n}$};
    \node[fac,above of=y] (f) {} edge (y);
    \node[obs,above of=f] (ty) {$\tilde{y}^{x_n} _{m}$} edge (f);
    \node[fac,above of=ty] (fp) {} edge (ty);
    \node[obs,left of=fp,xshift=-.25cm] (eta) {$\eta_{m} ^{x_n}$} edge[bend right] (f);
    \node[obs,above of=fp] (txi) {$\tilde{\xi}_{mk} ^{x_n}$} edge (fp);
    \node[fac,above of=txi] (fx) {} edge (txi) edge[bend right] (eta);
    \node[obs,above of=fx] (xi) {$\xi_{mk} ^{x_n}$} edge (fx);

    \node[var,right of=txi,xshift=.35cm] (w) {$w_{mk}$} edge (fp);

    \node[fac,right of=w,yshift=-0.75cm,xshift=.35cm] (fs) {} edge (w);
    \node[obs,above of=fs,yshift=1.5cm] (xis) {$\xi_{mk} ^*$} edge (fs);
    \node[obs,right of=fs,xshift=.35cm] (es) {$\eta^* _m$};
    \node[var,below of=fs] (tys) {$\tilde{y}^* _m$} edge (fs);
    \node[fac,below of=tys](fs) {} edge (tys) edge[bend right] (es);
    \node[var,below of=fs] (ys) {$y^*$} edge (fs);

    \node[dirprior] at (1.25,5.4) (lp) {};
    \node[fac] at (-1,5.4) {} edge (lp) edge(eta);
    \node[fac] at (3.3,5.4) {} edge(lp) edge(es);
    \node[anchor=south] at (lp.north) {$\lambda_m$};

    \node at (1.5,-1) {\sf\scriptsize locally weighted regression};
 \end{scope}
  \end{tikzpicture}

%% file: Experiments.tex
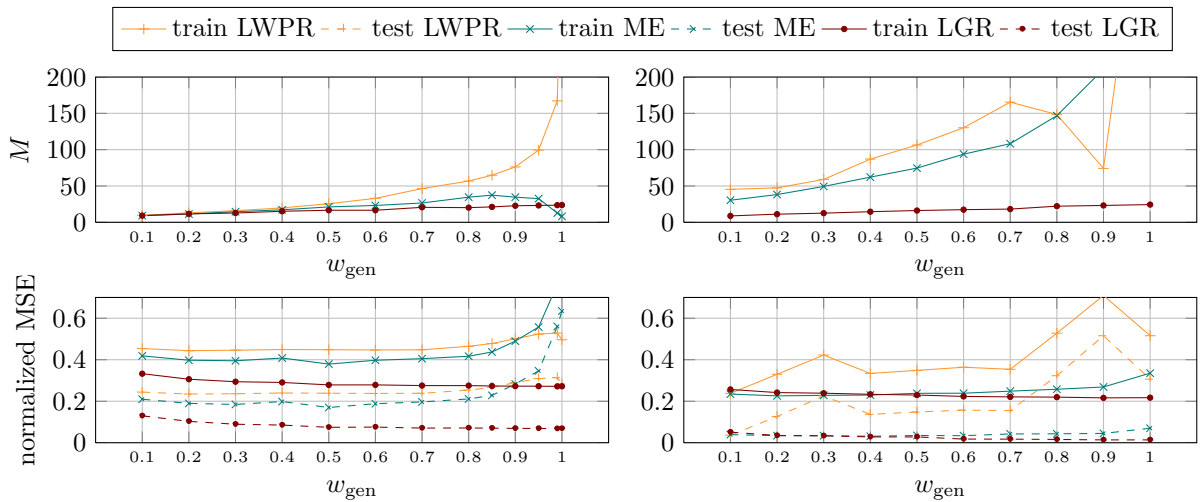
\begin{figure*}[htb]
\centering
   \beginpgfgraphicnamed{cross2D_results_barplot-external}%
   \input{cross2D_results_barplot.tikz}%
   \endpgfgraphicnamed%
 
\caption{\small{normalized mean squared error (nMSE) on 2D cross data without
  (left) and with (right) lengthscale learning.}}
\label{fig:cross_fun_res}
\end{figure*}
\begin{table*}[t]
\caption{Predictive Performance on SARCOS task}
\label{tab:SARCOS}
\begin{center}
\begin{tabular}{c |c c c|c c c|c c c}
  \toprule
  & & {\bf LWPR} & & &{\bf LGR} & &  &{\bf ME} & \\
\midrule
 & nMSE& (MSE) & $\#$ of LM & nMSE & (MSE) & $\#$ of LM &nMSE & (MSE) & $\#$ of LM \\ 
\midrule
J1   & $0.045$ & $(19.160)$  & 419 &$\mathbf{0.015}$ &
$\mathbf{(6.282)}$& 366 & 0.071 & (30.125) & 249 \\ 
J2    & $0.039$ & $(8.938)$   & 493 & $\mathbf{0.0195}$ & $\mathbf{(4.391)}$ & 361 &0.0871
& (19.564) & 257 \\ 
J3    & $0.034$ &$(3.420)$   & 483 &  $\mathbf{0.014}$ & $\mathbf{(1.382)}$  & 359 & 0.074&
(7.319) & 243  \\
J4   &  $0.024$ & $4.552$  & 384 & $\mathbf{ 0.016 }$  &
$\mathbf{(3.118)}$  & $\mathbf{348}$ & 0.039 & (7.348) & 239 \\
J5    & $0.064$& $(0.060)$  & 514 & $\mathbf{0.035}$ & $\mathbf{(0.0336)}$ & 354 & 0.058 &
(0.0552) & 438 \\ 
J6    & $0.075$ & $(0.221)$   & 519 & $\mathbf{0.027}$ & $\mathbf{(0.0786)}$ & 359 & 0.124 &
(0.363) & 234 \\ 
J7    & $0.03$ & $(0.203)$   & 405 &$\mathbf{0.023}$ &
$\mathbf{(0.1578)}$ & 358 & 0.0235& (0.1589) & 383 \\
\bottomrule
\end{tabular}
\end{center}
\vspace{-0.2cm}
\end{table*}
\section{Experiments}
We evaluate and compare to mixture of experts and locally weighted
projection regression (LWPR) -- an extension of LWR suitable for
regression in high dimensional space \cite{vijayakumar2000locally} --
on two data sets. The first is data drawn from the ``cross function''
in Figure~\ref{fig:cross_fun}, often used to demonstrate locally
weighted learning. For the second comparison we learn inverse dynamics
of a SARCOS anthropomorphic robot arm with seven degrees of freedom
\cite{RasmussenWilliams}.

In order to compare to mixture of experts we assume linear expert
models. A prior of the form $\N(\w_m;0, A_m^{-1})$ on each experts
regression parameters and normalized
gaussian kernels for the mixture components are used in our
implementation to make it as comparable as possible. To compute the
posterior in the {\sc e}-step a mean field approximation is employed.

In both experiments, LWPR performed multiple cycles through the data
sets. Both the local Gaussian regression and the mixture of experts
implementation are executed based on Algorithm \ref{alg:incLGR} and
are allowed an additional $1000$ iterations to reach convergence.
\subsection{Data from the `Cross Function'}
We used $2,000$ uniformly distributed training inputs, with zero mean
Gaussian noise of variance $(0.2)^2$. The test set is a regular grid
of $1641$ edges without noise and is used to evaluate how well the
underlying function is captured. The initial length scale was set to
$\lambda_m = 0.3, \forall m$, and we ran each method with and without
lengthscale learning (LSL) for $w_\text{gen}= 0.1,0.2, \dots,
1.0$. All results presented here are results averaged over 5 randomly
seeded runs.  Table~\ref{tab:cross_fun} presents the top performance
achieved by each method, including what the optimal setting for
$w_\text{gen}$ was and how many local models were used. In both
settings (with or without LSL) LGR outperforms both LWPR and ME in
terms of accuracy as well as number of local models used. To
understand the role of parameter $w_\text{gen}$ better, we also
summarize the normalized mean squared error as a function of parameter
$w_\text{gen}$ for all 3 methods in Figure~\ref{fig:cross_fun_res}
with (right) and without (left) LSL. The key message of this graph is
that the parameter $w_\text{gen}$ does not affect the performance of
LGR greatly. While accuracy slightly improves with increasing
$w_\text{gen}$ it is not dramatic. Thus for LGR $w_\text{gen}$ can be
thought of as a trade-off parameter, for smaller $w_\text{gen}$ the
algorithm has to consider less potential local models, for a slight
performance decrease. For LWPR and ME this relationship to
$w_\text{gen}$ is not that clear. Furthermore, although LGR has to
consider more local models for larger $w_\text{gen}$ (up to $2000$ for
$w_\text{gen}=1$), we only see a slight increase in the number of
local models in the final model, indicating that the pruning mechanism
works very well.

Finally, we show representative results of the shape of learned local
models for LWPR, LGR and ME In Figure ~\ref{fig:cross_fun}, nicely
illustrating the key difference between the three methods: In LWPR
local models don't know of each other and thus aim to find the best
linear approximation to the function. In both ME and LGR, the local
models know of each other and collaborate to fit the function.
\subsection{Inverse Dynamics Learning Task}
The SARCOS data contains $44,484$ training data points and $4,449$
test data points. The 21 input variables represent joint positions,
velocities and accelerations for the 7 joints. The task is to predict
the $7$ joint torques. In Table~\ref{tab:SARCOS} we show the
predictive performance of LWPR, ME and LGR when trained with
lengthscale learning and with $w_\text{gen}= 0.3$. LGR outperforms
LWPR and ME in terms of accuracy for almost all joints.
However, the true advantage of LGR lies in the fact
the number of hand tuned parameters is reduced to setting the learning
rate for the gradient descent updates and setting the parameter
$w_\text{gen}$.


%% file: cross2D_results_barplot.tikz
\usepgfplotslibrary{groupplots}
\begin{tikzpicture} 
\begin{groupplot}[group style={group size=2 by 2}]
 
\nextgroupplot[
        grid,
        xtick={0.1, 0.2, 0.3, 0.4, 0.5, 0.6, 0.7, 0.8, 0.9 , 1.0},
        xticklabel style={font=\tiny,rotate=0},
        xlabel= $w_\text{gen}$,
        ylabel = $M$,
        ymin = 0.0,
        ymax = 200,
        xmin = 0.0,
        xmax = 1.1,
        height=100pt,
        width=0.49\linewidth,
        enlarge x limits=0,
      ]
      \addplot[mark=+,ora] table[x index=0,y index=3]
                {cross2D_nolsl_nLM_2.txt};
      \addplot[mark=x,MPG] table[x index=0,y index=2]
                {cross2D_nolsl_nLM_2.txt};
      \addplot[mark=*,dred,mark options={scale=.5}] table[x index=0,y index=1]
               {cross2D_nolsl_nLM_2.txt};

\nextgroupplot[
        grid,
        xtick={0.1, 0.2, 0.3, 0.4, 0.5, 0.6, 0.7, 0.8, 0.9 , 1.0},
        xticklabel style={font=\tiny,rotate=0},
        xlabel= $w_\text{gen}$,
        ymin = 0.0,
        ymax =200,
        xmin = 0.0,
        xmax = 1.1,
        height=100pt,
        width=0.49\linewidth,
        enlarge x limits=0,
      ]
      \addplot[mark=+,ora] table[x index=0,y index=3]
                {cross2D_lsl_nLM_2.txt};
      \addplot[mark=x,MPG] table[x index=0,y index=2]
                {cross2D_lsl_nLM_2.txt};
      \addplot[mark=*,dred,mark options={scale=.5}] table[x index=0,y index=1]
               {cross2D_lsl_nLM_2.txt};

   \nextgroupplot[
        grid,
        xtick={0.1, 0.2, 0.3, 0.4, 0.5, 0.6, 0.7, 0.8, 0.9 , 1.0},
        xticklabel style={font=\tiny,rotate=0},
        ylabel= normalized MSE,
        xlabel= $w_\text{gen}$,
        ymin = 0.0,
        ymax = 0.7,
        xmin = 0.0,
        xmax = 1.1,
        height=100pt,
        width=0.49\linewidth,
        enlarge x limits=0,
        legend style={at={(2.1,3.0)},
          legend columns=-1},
      ]
      \addplot[mark=+,ora] table[x index=0,y index=3]
                {cross2D_nolsl_2.txt};
      \addplot[dashed,mark=+,ora] table[x index=0,y index=6]
               {cross2D_nolsl_2.txt};
      \addplot[mark=x,MPG] table[x index=0,y index=2]
                {cross2D_nolsl_2.txt};
      \addplot[dashed,mark=x,MPG] table[x index=0,y index=5]
               {cross2D_nolsl_2.txt};
      \addplot[mark=*,dred,mark options={scale=.5}] table[x index=0,y index=1] 
               {cross2D_nolsl_2.txt};
      \addplot[dashed,mark=*,dred,mark options={scale=.5}]  table[x index=0,y index=4]
                {cross2D_nolsl_2.txt};
      \legend{train LWPR, test LWPR, train ME, test ME, train LGR , test LGR}

\nextgroupplot[
        grid,
        xtick={0.1, 0.2, 0.3, 0.4, 0.5, 0.6, 0.7, 0.8, 0.9 , 1.0},
        xticklabel style={font=\tiny,rotate=0},
        xlabel= $w_\text{gen}$,
        ymin = 0.0,
        ymax = 0.7,
        xmin = 0.0,
        xmax = 1.1,
        height=100pt,
        width=0.49\linewidth,
        enlarge x limits=0,
      ]
      \addplot[mark=+,ora] table[x index=0,y index=3]
                {cross2D_lsl_2.txt};
      \addplot[dashed,mark=+,ora] table[x index=0,y index=6]
               {cross2D_lsl_2.txt};
      \addplot[mark=x,MPG] table[x index=0,y index=2]
                {cross2D_lsl_2.txt};
      \addplot[dashed,mark=x,MPG] table[x index=0,y index=5]
               {cross2D_lsl_2.txt};
      \addplot[mark=*,dred,mark options={scale=.5}] table[x index=0,y index=1]
               {cross2D_lsl_2.txt};
      \addplot[dashed,mark=*,dred,mark options={scale=.5}] table[x index=0,y index=4]
                {cross2D_lsl_2.txt};
 
\end{groupplot} 
 \end{tikzpicture}

%% file: paper.bbl
\begin{thebibliography}{10}

\bibitem{schaal1998constructive}
Stefan Schaal and Christopher~G Atkeson.
\newblock Constructive incremental learning from only local information.
\newblock {\em Neural Computation}, 10(8):2047--2084, 1998.

\bibitem{bottou1992local}
L{\'e}on Bottou and Vladimir Vapnik.
\newblock Local learning algorithms.
\newblock {\em Neural computation}, 4(6):888--900, 1992.

\bibitem{jacobs1991adaptive}
Robert~A Jacobs, Michael~I Jordan, Steven~J Nowlan, and Geoffrey~E Hinton.
\newblock Adaptive mixtures of local experts.
\newblock {\em Neural computation}, 3(1):79--87, 1991.

\bibitem{waterhouse1996bayesian}
Steve Waterhouse, David MacKay, and Tony Robinson.
\newblock Bayesian methods for mixtures of experts.
\newblock {\em Advances in neural information processing systems}, pages
  351--357, 1996.

\bibitem{hannah2011dirichlet}
Lauren Hannah, David~M Blei, and Warren~B Powell.
\newblock Dirichlet process mixtures of generalized linear models.
\newblock {\em Journal of Machine Learning Research}, 12:1923--1953, 2011.

\bibitem{Ting:2008uc}
Jo-Anne Ting, Mrinal Kalakrishnan, Sethu Vijayakumar, and Stefan Schaal.
\newblock {Bayesian Kernel Shaping for Learning Control}.
\newblock In {\em Neural information processing systems}, 2008.

\bibitem{Edakunni:2007uf}
Narayanan~U Edakunni, Stefan Schaal, and Sethu Vijayakumar.
\newblock Kernel carpentry for online regression using randomly varying
  coefficient model.
\newblock In {\em Proceedings of the international joint conference on
  artificial intelligence (IJCAI)}, 2007.

\bibitem{RasmussenWilliams}
C.E. Rasmussen and C.K.I. Williams.
\newblock {\em Gaussian Processes for Machine Learning}.
\newblock MIT Press, 2006.

\bibitem{micchelli2006universal}
C.A. Micchelli, Y.~Xu, and H.~Zhang.
\newblock Universal kernels.
\newblock {\em Journal of Machine Learning Research}, 7:2651--2667, 2006.

\bibitem{gibbs1997bayesian}
M.~N Gibbs.
\newblock {\em Bayesian Gaussian processes for regression and classification}.
\newblock PhD thesis, University of Cambridge, 1997.

\bibitem{rasmussen2002infinite}
C.E. Rasmussen and Z.~Ghahramani.
\newblock {Infinite mixtures of Gaussian process experts}.
\newblock In {\em Advances in neural information processing systems}. MIT
  Press, 2002.

\bibitem{neal1996bayesian}
R.M. Neal.
\newblock {\em {Bayesian learning for neural networks}}.
\newblock Springer Verlag, 1996.

\bibitem{Tipping:2001wb}
M.E. Tipping.
\newblock {Sparse Bayesian learning and the relevance vector machine}.
\newblock {\em The Journal of Machine Learning Research}, 1:211--244, 2001.

\bibitem{dsouza2004}
Aaron D'Souza, Sethu Vijayakumar, and Stefan Schaal.
\newblock The bayesian backfitting relevance vector machine.
\newblock In {\em Proceedings of the twenty-first international conference on
  Machine learning}, page~31. ACM, 2004.

\bibitem{titsias09:_variat_learn_induc_variab_spars_gauss_proces}
M.K. Titsias.
\newblock Variational learning of inducing variables in sparse {G}aussian
  processes.
\newblock In {\em Artificial Intelligence \& Statistics (AISTATS)}, volume~12,
  2009.

\bibitem{quinonero2002incremental}
Joaquin Quinonero-Candela and Ole Winther.
\newblock Incremental gaussian processes.
\newblock In {\em Advances in neural information processing systems}, pages
  1001--1008, 2002.

\bibitem{vijayakumar2000locally}
Sethu Vijayakumar and Stefan Schaal.
\newblock Locally weighted projection regression: An o (n) algorithm for
  incremental real time learning in high dimensional space.
\newblock In {\em Proceedings of the Seventeenth International Conference on
  Machine Learning (ICML 2000)}, volume~1, pages 288--293, 2000.

\end{thebibliography}
